%% file: _main.tex
\documentclass[runningheads]{llncs}
\usepackage{graphicx}
\usepackage{amsmath,amsfonts,amssymb}

\usepackage{float}
\usepackage{multirow,booktabs}
\usepackage{subfigure}
\usepackage[linesnumbered,ruled,vlined]{algorithm2e}
\usepackage{bm}
\usepackage{enumitem}
\usepackage{siunitx} 
\usepackage{diagbox}

\usepackage{hyperref}

\usepackage{wrapfig,lipsum,booktabs}
\usepackage{multicol}
\usepackage{caption} 
\usepackage{tikz}
\usetikzlibrary{positioning,calc,intersections}

\definecolor{forestgreen}{rgb}{0.13, 0.55, 0.13}
\definecolor{conceptor}{HTML}{996300}
\definecolor{sw1}{HTML}{996300}
\definecolor{sw0.8}{HTML}{b27300}
\definecolor{sw0.6}{HTML}{cc8400}
\definecolor{sw0.4}{HTML}{e59400}
\definecolor{sw0.2}{HTML}{ffa500}

\SetCommentSty{mycommfont}
\SetKwInput{KwInput}{Input}                
\SetKwInput{KwOutput}{Output}     

\usepackage[b]{esvect}
\renewcommand*{\vec}[1]{\vv{#1}}
\DeclareMathOperator{\mean}{mean}
\DeclareMathOperator{\std}{std}
\graphicspath{ {./imgs/} }

\begin{document}
\title{Joint Multiclass Debiasing of Word Embeddings}
\author{Radomir Popovi\'c\inst{1}  \and
Florian Lemmerich\inst{2} \and
Markus Strohmaier\inst{3}
}
\authorrunning{Popovi\'c et al.}
\institute{RWTH Aachen University, Germany$^{1,2,3}$\\ Gesis-Leibniz Institute for Social Sciences, Germany$^3$}

\maketitle 

\begin{abstract}
Bias in Word Embeddings has been a subject of recent interest, along with efforts for its reduction. Current approaches show promising progress towards debiasing single bias dimensions such as gender or race. In this paper, we present a joint multiclass debiasing approach that is capable of debiasing \emph{multiple} bias dimensions simultaneously. In that direction, we present two approaches, HardWEAT and SoftWEAT, that aim to reduce biases by minimizing the scores of the Word Embeddings Association Test (WEAT). We demonstrate the viability of our methods by debiasing Word Embeddings on three classes of biases (religion, gender and race) in three different publicly available word embeddings and show that our concepts can both reduce or even completely eliminate bias, while maintaining meaningful relationships between vectors in word embeddings. Our work strengthens the foundation for more unbiased neural representations of textual data.

\keywords{Word Embedding \and Bias Reduction \and Joint Debiasing \and WEAT.}
\end{abstract}

\input{sections/01_introduction.tex}
\input{sections/02_related_work.tex}
\input{sections/03_approach.tex}

\input{sections/04_experiments.tex}
\input{sections/05_discussion.tex}
\input{sections/06_conclusions.tex}

\section*{Acknowledgements}
We want to thank S. Karve, L. Ungar, L., and J. Sedoc for providing the code for Conceptor Debiasing and their support.  

% ---- Bibliography ----
\bibliographystyle{splncs04}
\bibliography{bib.bib}

\end{document}

%% file: sections/01_introduction.tex
\section{Introduction}
\label{sec:intro}
Word Embeddings, i.e., the vector representation of natural language words, are key components of many state-of-the art algorithms for a variety Natural Language Processing tasks, such as Sentiment Analysis or Part of Speech Tagging. 
Recent research established that popular embeddings are prone to substantial biases, e.g., with respect to gender or race ~\cite{DBLP:journals/corr/abs-1711-08412,bolukbasi2016}, which demonstrated in results like \emph{``Man is to Computer Programmer as Woman is to
Homemaker"}~\cite{bolukbasi2016} as results of basic analogy tasks.
Since such biases can potentially have an effect on downstream tasks, several relevant approaches for debiasing existing word embedding have been developed.
A common deficit of existing techniques is that debiasing is limited to a single bias dimension (such as gender). Thus, in this paper, we propose two new post-processing methods for joint/simultaneous multiclass debiasing, which differ in their trade-off between maintaining word relationships  and decreasing bias levels: HardWEAT completely eliminates contained bias as measured by the established Word Embedding Association Test \cite{caliskan2017}. SoftWEAT has a stronger and tunable emphasis on maintaining the original relationships between words in addition to bias removal. We demonstrate the effectiveness of our approach on the bias dimensions gender, race and religion on three conventional Word Embedding models: FastText, GloVe and Word2Vec. 

%% file: sections/02_related_work.tex
\section{Background and Related Work}
\label{sec:background}

In this section, we introduce key concepts formally and discuss related work on debiasing word embeddings.

We assume a \emph{Word Embedding} with vocabulary size $v$ that maps each word $w$ to a vector representation $\vec{w} \in \mathbf{R}^d$. Protected \emph{classes} $\operatorname{c} \in \operatorname{C}$ are entities on which bias can exist, e.g., race, religion, gender, or age. \emph{Subclasses} $\operatorname{S}$  or \emph{target set of words} refer to directions of a class, e.g., \emph{male, female} for $c=$\emph{gender}, and are associated with a set of definitional words $S_c$.
The set of \emph{neutral words} $\operatorname{N}$ contains all words that should not be associated with any $\operatorname{S_c} \, \forall \, c \in C$.
Finally, \emph{attribute sets of words} $\operatorname{A}$ ($\operatorname{A} \subset{\operatorname{N}}$) denote word sets that a target set of words can potentially be linked to, e.g., \emph{\{pleasant, nice, enjoyable\}} or \emph{\{science, research, academic\}}.

\noindent\textbf{The Word Embedding Association Test. }
\label{sec:weat}
The state-of-the art way of measuring biases in embeddings is the Word Embedding Association Test (WEAT) \cite{caliskan2017}:
It considers two sets of \emph{attribute words} ($A$, $B$), e.g., family and career related words, and two target sets ($X$, $Y$), e.g., black and white names.
The null hypothesis of this test states, that the relative association of target sets' words to both attribute sets' words are equally strong. 
Thus, rejecting the null hypothesis asserts bias.
To examine this, a test statistic $s$ quantifies how strongly $X$ is associated to $A$ in comparison to the association between $Y$ and $B$. 
It is computed as $s(X, Y, A, B) = \sum_{\vec{x} \in X} h(\vec{x}, A, B) - \sum_{\vec{y} \in Y} h(\vec{y}, A, B)$, where $h(\vec{w}, A, B) = \mean_{\vec{a} \in A}\cos(\vec{w}, \vec{a}) - \mean_{\vec{b} \in B}\cos(\vec{w}, \vec{b})$ describes the relative association between a single target word $x \in X$ compared to the two attribute sets in a range $[-2, 2]$.
Based on $s$, we assess the statistical significance via an one-sided permutation test through which we acquire $p$ value;
Additionally,  an effect size $d$ that quantifies the severity of the bias can be calculated as:
\begin{displaymath}
  d(X, Y, A, B) = \frac{\mean_{\vec{x} \in X} h(\vec{x}, A, B) - \mean_{\vec{y} \in Y} h(\vec{y}, A, B)}{\std_{w \in X \cup Y }h(\vec{w}, A, B)}
\end{displaymath}
We will use this effect size in our novel debiasing approaches.
\noindent\subsubsection{Existing debiasing techniques.}
To achieve reduction of bias, we will substantially extend the work of Bolukbasi et al. \cite{bolukbasi2016}, which describes two ways of debiasing Word Embeddings in a post-processing step: \textit{Hard and Soft Debiasing}. Both rely on identifying gender bias subspace $B$ via Principal Component Analysis computed on gender word pairs differences, such as \emph{he--she, and man--woman}. 
Hard debiasing employs a \textit{neutralize} operation that removes, e.g., all non-gender related words $\operatorname{N}$ from a gender subspace by deducting from vectors their bias subspace projection. Subsequently, an \textit{equalize} operation positions opposing gender pair words (e.g., king, queen) to share the same angle with neutral words. 
By contrast, Soft Debiasing enables more gradual bias removal by utilizing a tuning parameter $\lambda$: An embedding $W \in \mathbb{R}^{d \times |vocab|}$ is being transformed by optimizing a transformation matrix $T \in \mathbb{R}^{d \times d}$: \[\underset{T}{\text{min}} \| (TW)^T (TW) - W^T W\|^2_F + \lambda\| (TN)^T (TB) \|^2_F \] 

Another recent approach by Manzini et al. \cite{manzini2019black} incorporates these ideas, but expands and evaluates results not only on gender, but separately also on race and religion. It suggests a bias subspace definition for non-binary class environment, which is formulated via PCA of mean shifted $k$-tuples ($k=$number of subclasses) of the definitional words. 
There are also other recently proposed gender bias post-processing \cite{DBLP:journals/corr/abs-1901-07656,font2019equalizing} and pre-processing techniques \cite{DBLP:journals/corr/abs-1809-01496}.

In terms of existing joint multiclass debiasing techniques, \emph{Conceptor Debiasing} \cite{karve-etal-2019-conceptor}, is based on applying Boolean-like logic operators using soft transformation matrix on a linear subspace where Word Embedding has the highest variance. We will use this technique for comparison in our experiments. 

%% file: sections/03_approach.tex
\section{Approach}
\begin{figure}[t!]
\centering
\includegraphics[width=0.8\textwidth]{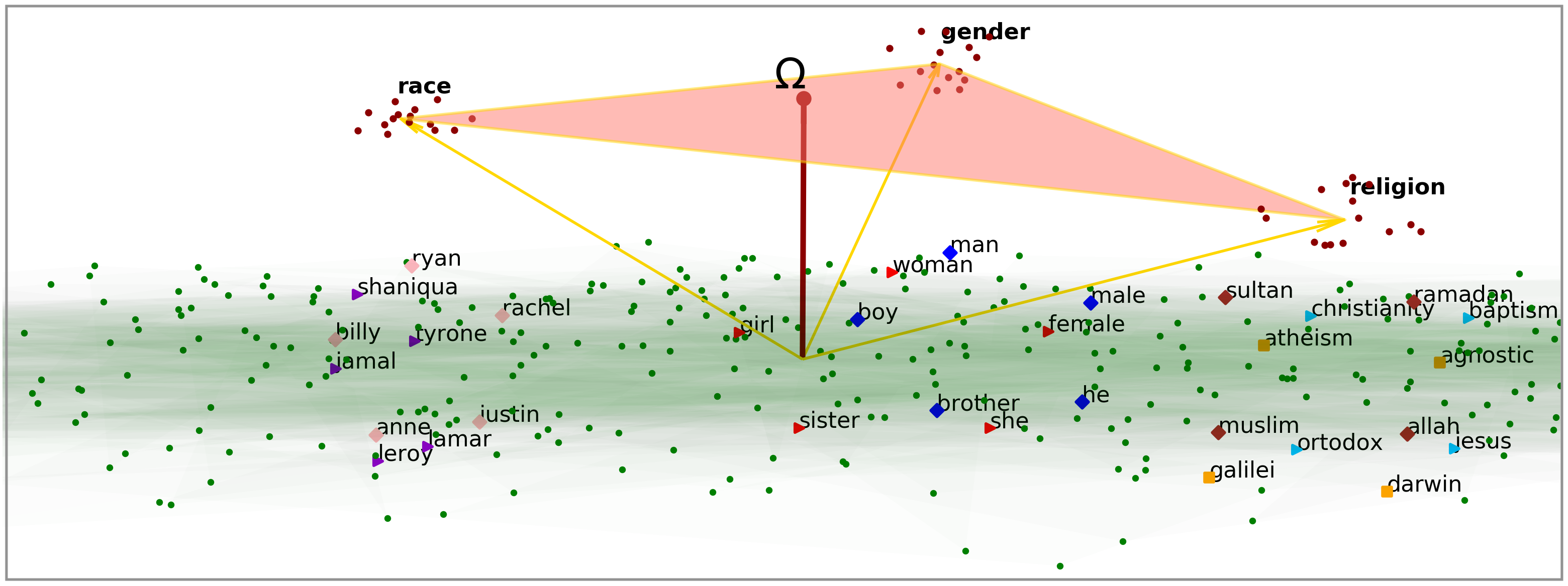}
\caption{Visual interpretation of multiclass debiasing via HardWEAT.}
\label{fig:hardweat_vis}
\end{figure}

In this chapter, we present our novel debiasing techniques. Our methods substantially extend previous works of Bolukbasi et al. \cite{bolukbasi2016} and Caliskan et al. \cite{caliskan2017}. 

\noindent\textbf{HardWEAT}: To adapt the \emph{neutralize} step from Bolubasi's work \cite{bolukbasi2016} jointly in a multiclass debiasing setting,  we first define the concept of \emph{class definitional vectors} $\vec{def_c}$ of classes (e.g., race, gender, \ldots), which are computed as the top component of a PCA on the vector representations of  definitional words $D_i \, \forall i \in \{1,...,n\}$ for $n$ subclasses (e.g., male, black, \ldots), cf. \cite{manzini2019black}. Now, each particular class can vary in terms of bias amount according to WEAT tests with chosen sets of attribute and target words.
Thus, we aggregate WEAT effect sizes $d$ for each class $c$ into bias levels  $\delta_c$ by averaging twice:
First, we compute the mean value for all target/subclass pairs within a class.
Second, we average those means for all results for class $c$.
Then, we compute a \emph{Centroid} $\Omega$, as the average of the class definitional vectors weighted by their normalized bias levels:
$\overrightarrow{\Omega} = \sum_{c \in C} \frac{(\delta_c \cdot \overrightarrow{def}_c)}{\sum_{c \in C} \delta_c}$.
We use this centroid to perform \emph{neutralization}, i.e., we re-embed all neutral words such that they perpendicular to it, cf.~\cite{bolukbasi2016}.

To adapt the \emph{equalize} step of Bolubasi's work, we move the definitional words of all subclasses in a way such that there is no relative preference of any attribute set towards a subclass.
For that purpose, we generate equally spread out points $e_1, \ldots, e_n$ on some circle with radius $r$ and center $\vec{o}$ as the new center points of the subclasses. In three or more dimensions, this circle can be defined by two vectors $(\vec{v_1}, \vec{v_2})$ perpendicular to  $\vec{o}$:
$
\overrightarrow{e_i} = \overrightarrow{o} +  r{\cos(\frac{2\pi i}{n})}\cdot\overrightarrow{v_1} + r{\sin(\frac{2\pi i}{n})}\cdot\overrightarrow{v_2}
$.

Given this formula, we use \emph{equidistancing} twice: First, we calculate new temporary central points for each class (e.g., gender) by neutralization (see above), then we determine new temporary central points $\vec{def_{S_{i}}}$ for each subclass (e.g., male) in a circle around that. Then, we compute new embeddings $\overrightarrow{w_i}$ for all definitional words $D_i$ in a circle around this subclass vector.
Note that $r_c, r_{S_c}$ are randomly generated with the condition $r_{S_c} \gg r_c$, whereas $\vec{v_1}, \vec{v_2}$ are obtained via SVD. 
Along with an illustration (see Figure \ref{fig:hardweat_vis}), we present the full procedure in Algorithm 1.

\begin{algorithm}[b!]
\caption{HardWEAT algorithm.}
\label{algorithm:hardweat}
\DontPrintSemicolon
\KwInput{Word Embedding matrix $\boldsymbol{W_{v \times d}}$, Words $W$, set of classes $C$, set of subclasses $S$, Bias levels $\delta_c$}
  \For(\tcp*[f]{Generating class definitional vectors}){$c \in C$} 
  { 
    $\vec{def_{c}} :=\operatorname{PCA} \Bigg(\bigcup\limits_{i=1}^{n}\bigcup\limits_{\vec{w} \in D_i} \vec{w} - \mu_i\Bigg)$ %\\
  }
  $\overrightarrow{\Omega} = \sum\limits_{c \in C} (\delta_c \cdot \overrightarrow{def}_c)$ \tcp*[f]{Computing centroid}
  
 $N = W \setminus B$\tcp*[f]{Defining neutral words}
 
    $\boldsymbol{N} := \boldsymbol{N} - \frac{\boldsymbol{N}\cdot \vec{\Omega}}{||{\overrightarrow{\Omega}}||}\overrightarrow{\Omega}$\tcp*[f]{Neutralizing}
    
    $\boldsymbol{W} = \frac{\boldsymbol{W} }{||\boldsymbol{W} ||}$\tcp*[f]{Normalizing vectors}
    
  \For(\tcp*[f]{Equidistancing}){$c \in C$}    
  { 
    $\vec{O_{c}} :=  \vec{def_{c}} - \frac{ \vec{def_{c}}\cdot \vec{\Omega}}{||{\overrightarrow{\Omega}}||}\overrightarrow{\Omega}$\\
  \For{$i \in \{1, ... , |S_c|\}$}     
  { 
 $\vec{def_{S_{i}}}\prime = \vec{O_{c}} +  r_c{\cos(\frac{2\pi i}{|S_i|})}\cdot\overrightarrow{v_1} + r_c{\sin(\frac{2\pi i}{|S_i|})}\cdot\overrightarrow{v_2}$\\
      \For{$w_j \in W_S$}
      {
      $\overrightarrow{w_j}
      = \vec{def_{S}}\prime +  r_{S_c}{\cos(\frac{2\pi j}{|W_S|})}\cdot\overrightarrow{v_1} + r_{S_c}{\sin(\frac{2\pi j}{|W_S|})}\cdot\overrightarrow{v_2}$\\
      }
 }
  }
$\boldsymbol{W} = \frac{\boldsymbol{W} }{||\boldsymbol{W} ||}$\tcp*[f]{Normalizing vectors}
\end{algorithm} 

For successful equidistancing, i.e., equal dispersion of words, we also require that the length of all subclass/target words within a particular class must be equal. 
Violations of this requirement will result in inadequate WEAT scores.
Furthermore, HardWEAT could in theory result in some equidistanced words becoming angle-close to random words: Thus, to counter this, we design this process in an iterative manner, modifying $r_{S_{c}}$ until there are no neutral words having an angle greater than certain threshold (e.g., we use $\ang{45}$ as a default). 
We discuss factors of success and open issues in more detail in Section \ref{sec:discussion}. 
Overall, the HardWEAT method ultimately aims at complete bias elimination, but randomizes the topological structure of subclass words.

\noindent\textbf{SoftWEAT} is a more gradual debiasing alternative with a greater focus on quality preservation. It provides the user with a choice on how to debias and to what extent.
Let us assume that a particular target set of words $S_{c}$  is closely related in terms of their angle to some attribute set of words $A_i$ (e.g., $A_1$ is the set of \emph{pleasant}  words, $A_2$ is the set of \emph{intellectual} words). 
Hypothetically, bias would be minimized if the subclass words $S_{c}$ would be perpendicular to the attribute sets $A_i$. 
Thus, moving $S_{c}$ towards such a perpendicular space/vector (null vector), noted as $\vec{n}$, results in bias reduction. 
Since full convergence towards a vector $\vec{n}$ may result in quality decrease we define a parameter $\lambda \in [0,1]$  as a \emph{level of removal}. 
Setting $\lambda = 1$ results in maximal angle decrease between $S_c$ and $A_i$ (perpendicular vectors), while $\lambda = 0$ makes no transformation at all. 
Note, however, that placing vectors from some subclass words $S_i$ to be perpendicular with attribute words $A_i$ may not necessarily result in overall bias level reductions since WEAT tests are relative measures. 
E.g., they take the relationship between \emph{male} and \emph{intellectual} words and the relationship between \emph{female} and \emph{appearance}-related words into account at the same time.
Additionally, higher $\lambda$ also pose a greater risk of producing new bias. Moving the representation of subclass words away from attribute words $A_1$ may result in moving them closer to some other $A_2$ without prior intent. For example, moving \emph{male} away from \emph{science} can get them closer to \emph{aggressive}, resulting in other WEAT test $d$ increase. 
To address these issues, we choose a nullspace via SVD  that minimizes the average WEAT effect size for all tests. 

SoftWEAT executes the following steps: 
Given some $S_c$ and its corresponding related attribute set of words $A_i, \ldots, A_n$, we generate a matrix $\boldsymbol{A_{n \times d}}$, where the $n$ rows consist of the first principal components for each $A_i$ respectively. 
For $\boldsymbol{A}$, we then find the nullspace vector $\vec{n}$ that decreases WEAT test scores the most. 
As a goal, we aim to translate our $S_c$ to this space. 
Since initially, there might be only few target set words $S_c$, we expand it by adding for all of them the closest, frequently occurring neighbor words (e.g., the 20 closest neighbors with minimum frequency of 200 in the English Wikipedia \cite{arora2017asimple}).)
Afterwards, we calculate a transformation vector $\vec{\psi}$ for which it holds that: $\vec{\psi} = \vec{n} - \vec{m}$, where $\vec{m}$ is the mean of $S_c$ words. 
This transformation vector is then multiplied with a parameter $\lambda \in [0,1]$ followed by conversion into linear translation matrix $\Psi$. 
By translating vectors from extended word sets, we preserve relationships of subclass words $S_c$ to these words.
Note that we only modify positions of words from the extended neighborhood of subclass words, but not all words from the vocabulary. Finally, as a last optional step, we normalize all vectors. 

In SoftWEAT, the user can decide on the $\lambda$, but can also select target and attribute sets used for debiasing. 
By default, we will only take those into account, for which WEAT scores result in  an aggregated effect size of $|d|>0.6$. To compute this, we accumulate attribute sets per each target set that form matrix \textbf{A} per each $S_c$. In case of positive $d$, $S_1$ gets removed from $A_1$, and $S_2$ from $A_2$. In case of negative $d$, removal is done other way around. 
The algebraic operations are formalized as follows:
Out of the new matrix $\textbf{W'}$ with $|S_c|$ rows and $d+1$ columns, all but the last row are taken as a new vector representation for the given target set of words $S_c$. 

\begin{gather*}
\quad\textbf{A}\vec{n} = 0 \quad\quad \vec{\psi} = \lambda(\vec{n} - \vec{m})\\
\boldsymbol{W_{S_c}'} = 
\Psi\left[\begin{array}{c}(W_{S_c})^T \\\hline1 \end{array}\right] = 
\begin{bmatrix}
1 & 0 & 0 & \hdots &\psi_1 \\
\vdots & \vdots & \ddots & \hdots & \vdots\\
0 & 0 & \hdots & 1 & \psi_d\\
0 & 0 & 0 &\hdots & 1
\end{bmatrix}\begin{bmatrix}
w_{11} & w_{21} & \hdots & w_{n1}  \\
\vdots & \vdots & \vdots & \ddots \\
w_{1d} & w_{2d} & \hdots & w_{nd} \\
1 & 1 & \hdots &1\\
\end{bmatrix}
\end{gather*}

%% file: sections/04_experiments.tex
\section{Experimental Evaluation}
\label{section:experiments}
This section presents example results of our methods in practical settings and compares it with Conceptor Debiasing \cite{karve-etal-2019-conceptor} as the current state-of-the-art approach. We measure the bias decrease along with deterioration of embedding quality and show the effects of biased/debiased embeddings in a Sentiment Analysis as a downstream task example. Due to limited space, we also provide an extended set of results in an accompanying online appendix\footnote{\url{https://git.io/JvL10}}.

\subsection{Experimental Setup}
\noindent\textbf{Datasets:} Extending the experimental design from \cite{10.1145/3342220.3343658}, we apply debiasing simultaneously on following target sets/subclasses: \emph{(male, female) - gender}, \emph{ (islam, christianity, atheism) - religion} and \emph{(black and white names) - race} with seven distinct attribute set pairs\footnotemark[5].
We collected target, attribute sets, and class definitional sets from literature \cite{may2019measuring,DBLP:journals/corr/abs-1809-01496,DBLP:journals/corr/abs-1804-06876,caliskan2017,manzini2019black,10.1145/3342220.3343658}, see our online appendix for a complete list.
As in previous studies \cite{karve-etal-2019-conceptor}, evaluation was done on three pretrained Word Embedding models with vector dimension of 300: FastText\footnote{\url{https://fasttext.cc/docs/en/english-vectors.html}}(English webcrawl and Wikipedia, $~$2 million words), GloVe\footnote{\url{https://nlp.stanford.edu/projects/glove/}}(Common Crawl, Wikipedia and Gigaword, $~$2.2 million words) and Word2Vec\footnote{\url{https://drive.google.com/uc?id=0B7XkCwpI5KDYNlNUTTlSS21pQmM}} (Trained on Google News, 3 million words). For the Sentiment Analysis task, we additionally employed a dataset of movie reviews \cite{maas-EtAl:2011:ACL-HLT2011}. 

\noindent\textbf{Quality Assessment}: First, we compared
ranked lists of word pairs in terms of their vector similarity to human judgement via Spearman’s Rank correlation coefficient \cite{doi:10.1002/0471667196.ess5050.pub2}, by using the collection taken from the Conceptor Debiasing evaluation \cite{karve-etal-2019-conceptor}, i.e., \emph{RG65, WS, RW, MEN, MTurk, SimLex, SimVerb}. In addition, we also utilize Mikolov Analogy Test \cite{mikolov2013a}.

\noindent\textbf{Methods}: Regarding HardWEAT, we specified neutral words through set difference between all words and ones from priorly defined target sets. In terms of SoftWEAT we provide details such as target-attribute sets structure and number of changed words in section F of online appendix. We applied the OR operator in the Conceptor Debiasing, using the same word set of subclasses within the three above defined classes (subspaces in Conceptor Debiasing).

\subsection{Bias levels and quality of Word Embeddings}
First, we focus on overall bias levels, see Table \ref{table:bias_lev},
by measuring WEAT scores before and after debiasing.
We observe that HardWEAT removes the measured bias completely as indicated by zero WEAT scores. 
SoftWEAT also substantially reduces the bias measurements to different degrees for different datasets. 
In comparison, for example with $\lambda=1$, SoftWEAT still leads to a stronger reduction in bias compared to the state-of-the-art Conceptor algorithm in all but one measurement.

\begin{table}[t] \centering
\begin{tabular}{|c|c|c|c|c|c|c|c|c|c|}  
\cline{3-10}

\multicolumn{1}{c}{} &
& {\multirow{2}{*}{REG}}
& {\multirow{2}{*}{\color{forestgreen}HW}} 
& {\color{sw0.2}SW} 
& {\color{sw0.4}SW}
& {\color{sw0.6}SW} 
& {\color{sw0.8}SW} 
& {\color{sw1}SW}
& {\multirow{2}{*}{\color{blue}CPT}} \\

\multicolumn{1}{c}{} & & & &
{\color{sw0.2}$\lambda=0.2$} 
& {\color{sw0.4} $\lambda=0.4$}
& {\color{sw0.6}$\lambda=0.6$} 
& {\color{sw0.8}$\lambda=0.8$} 
& {\color{sw1}$\lambda=1$}
&

 \\\cline{3-9}\hline 
& gender & 0.62 & 0.0** & 0.47 & 0.3 & 0.28 & 0.15 & 0.09* & 0.24\\ 
GloVe & race & 0.71 & 0.0** & 0.6 & 0.51 & 0.39 & 0.3 & 0.19* & 0.43\\ 
& religion & 0.77 & 0.0** & 0.62 & 0.46 & 0.29 & 0.2 & 0.16* & 0.28
\\\hline\hline
& gender & 0.52 & 0.0** & 0.14* & 0.17 & 0.32 & 0.31 & 0.32 & 0.46\\ 
FastText & race & 0.36 & 0.0** & 0.13* & 0.16 & 0.25 & 0.3 & 0.31 & 0.31\\ 
& religion & 0.6 & 0.0** & 0.27 & 0.21* & 0.29 & 0.35 & 0.42 & 0.54
\\\hline\hline
& gender & 0.63 & 0.0** & 0.48 & 0.37 & 0.32 & 0.2 & 0.23 & 0.12*\\ 
Word2Vec & race & 0.56 & 0.0** & 0.32 & 0.28 & 0.2 & 0.16 & 0.16* & 0.38\\ 
& religion & 0.47 & 0.0** & 0.31 & 0.19 & 0.11* & 0.18 & 0.2 & 0.28\\
\hline
\end{tabular}
\caption{Bias levels for gender, race, and religion before debiasing (REG) and after debiasing with Hard/SoftWEAT (HW/SW) or Conceptor (CPT)}

\label{table:bias_lev} 
\end{table}

\noindent Regarding quality assessment, Table \ref{fig:qual_glove},
shows the complete results for seven  rank similarity datasets as well as the Mikolov analogy score on the example of the GloVe Word Embedding.
As expected, a significant quality drop appears with HardWEAT, most notably on the RG65 dataset. With SoftWEAT, greater $\lambda$ induce larger modifications of the original embeddings, which corresponds to a greater drop in embedding quality. 
However, in three out of eight test settings, SoftWEAT with lower $\lambda$ settings achieved a similar or higher score and also leads to competitive results compared to Conceptor, see Section \ref{sec:discussion} for an in-depth discussion. Extended results can be found in the online appendix.

\begin{table*}[t!]\centering\begin{tabular}{|c||c|cccccc|c|}
\cline{2-9}
\multicolumn{1}{c|}{}
&
\multirow{2}{*}{REG}& 
\multirow{2}{*}{\color{forestgreen}HW}& 
{\color{sw0.2}SW} & 
{\color{sw0.4}SW}& 
{\color{sw0.6}SW} & 
{\color{sw0.8}SW}&
{\color{sw1}SW} &  
\multirow{2}{*}{\color{blue}CPT} \\ 
\multicolumn{1}{c|}{}
&
& 
& 
{\color{sw0.2}$\lambda=0.2$} & 
{\color{sw0.4}$\lambda=0.4$}& 
{\color{sw0.6}$\lambda=0.6$} & 
{\color{sw0.8}$\lambda=0.8$}&
{\color{sw1}$\lambda=1$} & 
\\\cline{2-9}\hline

RG65 & \textbf{76.03}* & 63.42 & 75.68 & 75.39 & 74.34 & 73.25 & 71.42 & 68.73 \\
WS & 73.8 & 69.73 & \textbf{74.0}* & 73.96 & 73.15 & 71.95 & 70.34 & 73.48 \\
RW & 46.15 & 46.09 & 46.15 & 46.13 & 46.08 & 46.02 & 45.94 & \textbf{52.45}* \\
MEN & 80.13 & 77.52 & 80.34 & \textbf{80.34}* & 80.1 & 79.57 & 78.74 & 79.99 \\
MTurk & \textbf{71.51}* & 69.78 & 71.25 & 70.78 & 70.37 & 69.48 & 68.64 & 68.24 \\
SimLex & 40.88 & 40.44 & 41.46 & 42.0 & 42.07 & 42.21 & 42.18 & \textbf{47.36}* \\
SimVerb & 28.74 & 28.74 & 28.9 & 29.15 & 29.27 & 29.54 & 29.78 & \textbf{36.78}* \\
\hline\hline
Mikolov & 0.65 & 0.64 & \textbf{0.65}* & 0.65 & 0.65 & 0.64 & 0.64 & 0.63 \\\hline
\end{tabular}
\caption{Measurements of quality tasks after debiasing for  GloVe embeddings.}
\label{fig:qual_glove}
\end{table*}

\subsection{Sentiment Analysis} To analyze the effects of debiasing in downstream tasks, we study a sentiment analysis task in the context of movie reviews, i.e., we investigate whether we observe significant differences in predicted sentiments when using biased and debiased Word Embeddings. 
Modifying setup from Packer et al. \cite{47172}, we added to the end of each test sentences randomized word from opposing set pairs (e.g., a typical black name or a typical white name), and calculated the difference in the prediction, i.e., the \emph{polarity score} according to a simple neural network that takes the respective pretrained word embeddings as pretrained embedding, which is followed by Flatten-operation and a sigmoid layer. The first two layers were not trainable.

\begin{wrapfigure}[15]{r}{0.55\textwidth}
\centering
\includegraphics[width=0.525\textwidth]{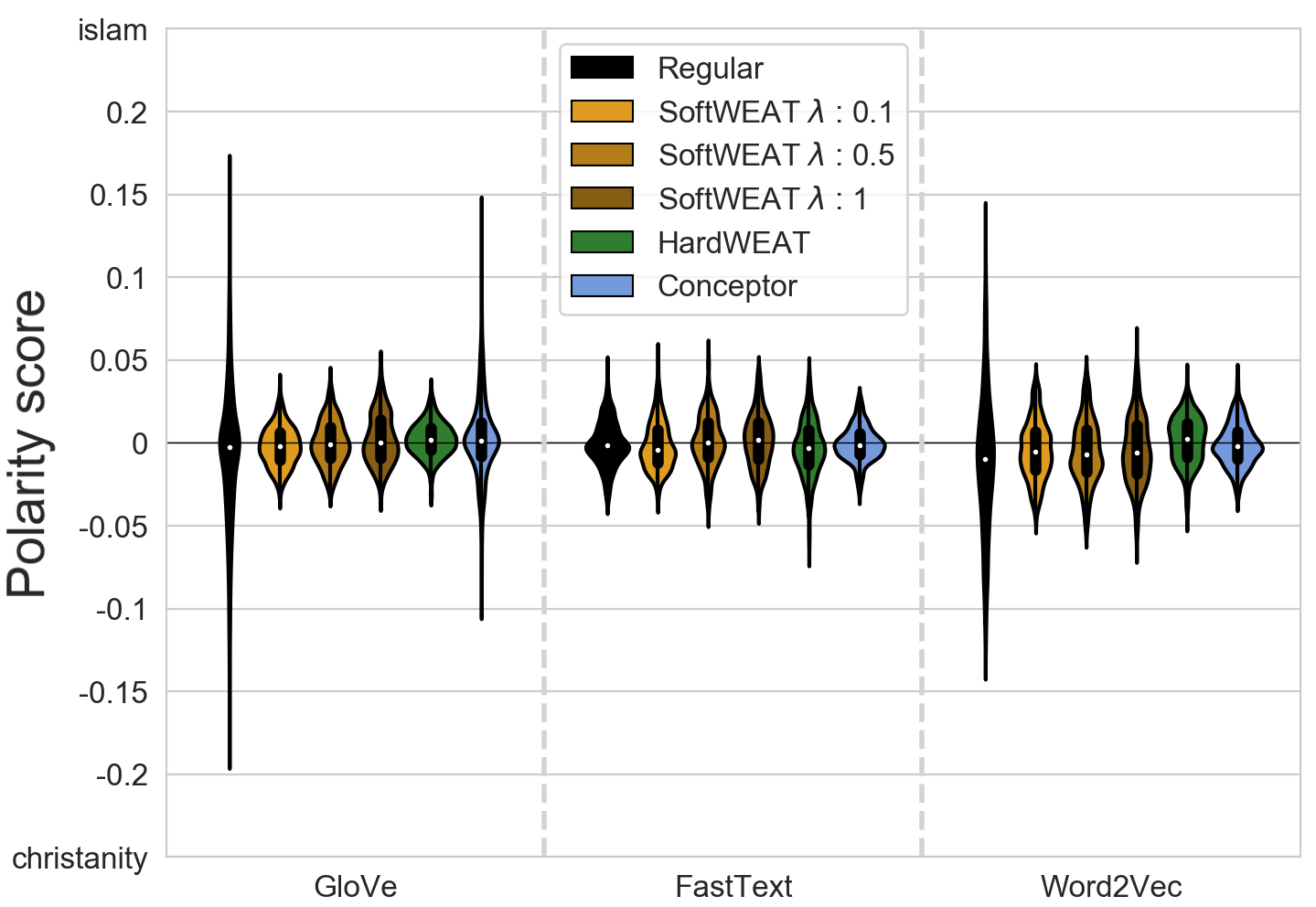}
\caption{Classifier Bias on (islam, christianity)}
\label{fig:class_bias}
\end{wrapfigure}

For training, we used a combination of binary-cross entropy loss and Ada-Delta optimizer. We trained our model only on sentences with a maximum length of 50 that did not contain any target sets of words. Shuffling test and training set, we then calculated \emph{polarity score} on 15 different model instances using 6 different kinds of Embeddings: Regular (not debiased), SoftWEAT with $\lambda=0.1, 0.5, 1$, HardWEAT and Conceptor Debiasing with $\alpha=2$. 

Results for \emph{[christianity, islam]} are shown in Figure \ref{fig:class_bias}.
For the non debiased embeddings, we observe that our classifier has no clear trend such that the addition of a bias word in a sentence influences the polarity in a particular direction. However, we can see that bias words have a strong influence on the classifier, leading to a large variance of \emph{polarity scores} in GloVe and Word2Vec-based models.
When we apply our debiasing methods, we see that the respective variances are substantially decreased, e.g., around 5.45 times for SoftWEAT already with a small $\lambda$  of $0.1$ in GloVe. This is only the case with Conceptor Debiasing for the Word2Vec-based model, but not for the GloVe-based model.

%% file: sections/05_discussion.tex
\subsection{Discussion}
\label{sec:discussion}

\textbf{HardWEAT and SoftWEAT success factors:} Based on our experiments, see also our online appendix, we could identify a variety of success factors for our methods. 
Regarding HardWEAT, higher embedding dimensionality and more dispersed vectors are more likely to output more desirable outcome. 
This is due to the possibility of random words appearing as close neighbors to a target words during the equidistancing procedure, which we counter iteratively as described above. 
Regarding centroid ($\overrightarrow{\Omega}$) neutralization, relevant factors of importance are number of classes, uniformity of bias levels and angles between all $def_c$: Having more classes with uniform bias levels and distant angles can output non-desirable results, e.g., too large distances between $\overrightarrow{\Omega}$ and $\vec{def_c}$ points. However, the user could also manually define an alternative centroid $\overrightarrow{\Omega}$. 
We acknowledge that \emph{equidistancing} comes with a main drawback, which is the partial loss of relationship between target and non-target words, which is also reflected in a quality drop according to different metrics.
Regarding SoftWEAT, decreased angles between target and attribute sets after translation may not necessarily result in lower bias levels, since WEAT is a relative measure. 
However, we take one nullspace which minimizes these values. 
Furthermore, we note that in our experiments we used all target and attribute set pairs within each of the WEAT tests, which can be further optimized: We may not want to debias something, which isn't priorly biased.
E.g., removing \emph{male} from \emph{science} may be necessary, whereas doing the same for \emph{female} from \emph{art} may not, thus we could exclude this latter pair.
Also, we should bear in mind, that attribute sets of words are often correlated (as also shown by our experiments in the online appendix). This implies that by moving subclass words towards a specific set of attribute words, we automatically change the associations also with other attribute sets. Thus, the user plays a crucial role in deciding which sets should be used for debiasing.

\textbf{Comparison with Conceptor Debiasing:} Given 
the experimental results, we conclude that neither Conceptor Debiasing nor SoftWEAT outperform each other. Yet, SoftWEAT exhibits some distinctive advantages: (i) With SoftWEAT, relationships within the target set words remain completely the same, whereas in Conceptor Debiasing, overall angle distribution gets more narrow (See online appendix for more details). (ii) We argue that with SoftWEAT, user gets more freedom with choosing on which target/attribute set combination and to which degree debiasing is applied. (iii) Given our method, there is no difference in word representation if one uses small subset of neutral words or complete vocabularies. Nevertheless, we acknowledge that Conceptor Debiasing does succeed in reducing bias equally well by applying it in more global behavior.

%% file: sections/06_conclusions.tex
\section{Conclusion}
In this paper, we presented two novel approaches for multi-class debiasing of Word Embeddings. We demonstrated the general viability of these methods for reducing and/or eliminating biases while preserving meaningful relationships of the original vector representations.
We also analyzed the effects of debiased representations in Sentiment Analysis as an example downstream task and find that debiasing leads to substantially decreased variance in the predicted polarity.
Overall, our work contributes to ongoing efforts towards providing more unbiased neural representations of textual data.